\documentclass[journal]{IEEEtran}
\usepackage{amsmath,amsfonts}
\usepackage{algorithmic}
\usepackage{algorithm}
\usepackage{array}
\usepackage[caption=false,font=footnotesize,labelfont=rm,textfont=rm]{subfig}
\usepackage{textcomp}
\usepackage{stfloats}
\usepackage{url}
\usepackage{verbatim}
\usepackage{graphicx}
\usepackage{amssymb}
\usepackage{amsmath}
\usepackage{xcolor}
\usepackage{booktabs}
\usepackage{threeparttable}
\usepackage{multirow}
\usepackage[colorlinks]{hyperref}

\begin{document}

\title{Affinity Contrastive Learning for Skeleton-based Human Activity Understanding}

\author{
Hongda Liu, Yunfan Liu, Min Ren, Lin Sui, Yunlong Wang, Zhenan Sun,~\IEEEmembership{Senior Member,~IEEE}
\thanks{
This work was supported in part by the National Natural Science Foundation of China under Grant U23B2054, Grant 62276263, Grant 62406304, and Grant 62406028, and in part by Tianjin Key Research and Development Program Chinese Academy of Sciences (CAS)-Cooperation Project under Grant 24YFYSHZ00290. (Corresponding author: Zhenan Sun.)
}
\thanks{
H. Liu, Y. Wang and Z. Sun are with the Institute of Automation, Chinese Academy of Sciences, Beijing 100190, China, and H. Liu is also with the School of Artificial Intelligence, University of Chinese Academy of Sciences, Beijing 100049, China (e-mail: hongda.liu@cripac.ia.ac.cn; yunlong.wang@cripac.ia.ac.cn; znsun@nlpr.ia.ac.cn).
}
\thanks{
Y. Liu is with the School of Electronic, Electrical and Communication Engineering, Univeristy of Chinese Academy of Sciences, Beijing 101408, China (e-mail: liuyunfan@ucas.ac.cn).
}
\thanks{
M. Ren is with the School of Artificial Intelligence, Beijing University of Posts and Telecommunications, Beijing 100876, China (e-mail: minren@bupt.edu.cn).
}
\thanks{
L. Sui is with the Moonshot AI, Beijing 100086, China (e-mail: suilin0432@gmail.com).
}
}

\markboth{IEEE Transactions on Biometrics, Behavior, and Identity Science}%
{Shell \MakeLowercase{\textit{et al.}}: Bare Demo of IEEEtran.cls for IEEE Journals}

\maketitle

\begin{abstract}
In skeleton-based human activity understanding, existing methods often adopt the contrastive learning paradigm to construct a discriminative feature space.
However, many of these approaches fail to exploit the structural inter-class similarities and overlook the impact of anomalous positive samples.
In this study, we introduce ACLNet, an \underline{A}ffinity \underline{C}ontrastive \underline{L}earning \underline{Net}work that explores the intricate clustering relationships among human activity classes to improve feature discrimination. 
Specifically, we propose an affinity metric to refine similarity measurements, thereby forming activity superclasses that provide more informative contrastive signals.
A dynamic temperature schedule is also introduced to adaptively adjust the penalty strength for various superclasses.
In addition, we employ a margin-based contrastive strategy to improve the separation of hard positive and negative samples within classes.
Extensive experiments on NTU RGB+D 60, NTU RGB+D 120, Kinetics-Skeleton, PKU-MMD, FineGYM, and CASIA-B demonstrate the superiority of our method in skeleton-based action recognition, gait recognition, and person re-identification.
The source code is available at \url{https://github.com/firework8/ACLNet}.
\end{abstract}


\IEEEpeerreviewmaketitle

\section{Introduction}

\IEEEPARstart{I}{n} recent years, skeleton-based human activity understanding has garnered significant research attention owing to its robustness under complex environmental conditions and high computational efficiency~\cite{song2018spatio,rao2023transg,liu2025revealing}.
Despite these advances, skeleton-based representations often suffer from inherent ambiguity when discriminating between visually similar activities, due to the absence of interacting objects (e.g., \textit{reading} vs. \textit{writing}) and detailed body shapes (e.g., \textit{waving} vs. \textit{making ok sign}).
This limitation becomes especially critical in biometric applications, where subtle behavioral cues are essential for activity characterization and identity inference.

Therefore, recent methods have turned to discriminative contrastive learning techniques~\cite{huang2023graph,zhou2023learning,li2025joint}.
These approaches either exploit global contextual cues across all skeleton sequences~\cite{huang2023graph} or decouple spatial and temporal features for contrastive refinement~\cite{zhou2023learning}.
By integrating contrastive constraints into the frameworks, they enhance the discriminability of skeleton representations.
However, there are two problems with the existing contrastive learning paradigms.

\begin{figure}[t]
\centering
\includegraphics[width=\linewidth]{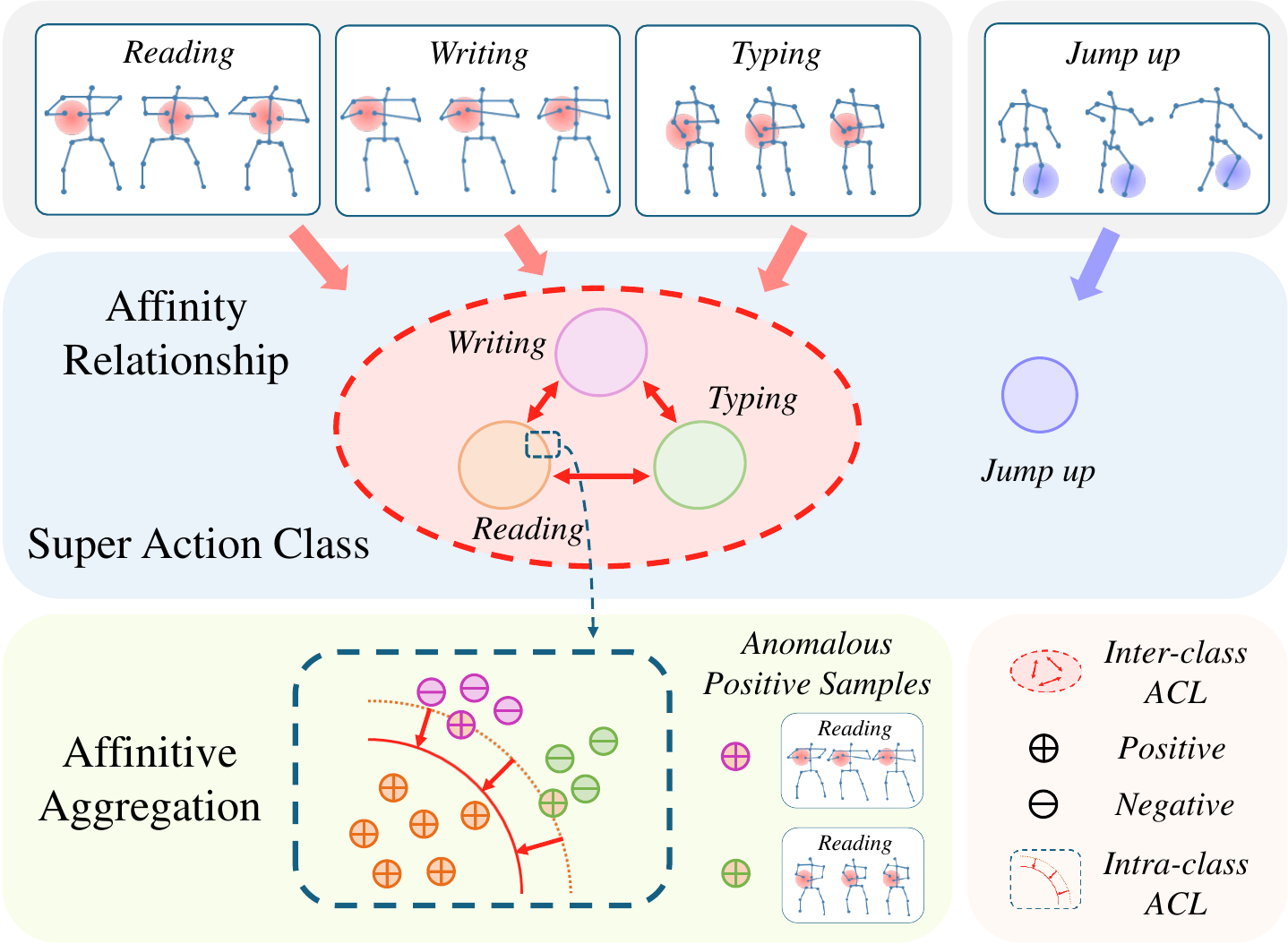}
\caption{ 
Conceptual diagram for Affinity Contrastive Learning.
The neglect of structural commonalities among classes and inherently anomalous positive samples within classes will degrade the performance of existing methods.
Therefore, we propose Affinity Contrastive Learning to improve discriminative representations at the inter-class and intra-class levels.
}
\label{fig:figure1}
\end{figure}

First, these methods fail to exploit the potential structural similarities between activity classes.
Intuitively, activities with similar motion patterns are prone to misclassification due to the commonalities in their skeleton sequences, such as common key joints or trajectories.
These physical similarities are then projected into the embedding space, potentially forming clusters of similar activities across different classes. 
The relationships reflected in these clusters provide rich supervisory signals for contrastive learning in the embedding space. 
However, existing methods rely solely on global positive-negative comparisons, ignoring the exploration of the structural information.
This leads to inefficient optimization, thereby limiting discriminative power in fine-grained scenarios.

Moreover, current methods neglect the inherent impact of anomalous positive samples within classes.
Intra-class variability, such as differences in observation angles and movement amplitudes, inevitably introduces noise.
This results in the possibility of some hard positives that are easily confused with samples from other classes.
The inherent impact of these anomalous positives, coupled with the disturbance caused by negative samples from hard classes, may lead to accumulated errors in the embedding space, ultimately degrading overall performance. 
This fact suggests that the learning manner for hard samples should be reformulated, however, which is not supported in the current paradigm.

To address the above issues, we introduce ACLNet, a novel affinity contrastive learning network for skeleton-based human activity understanding.
First, we propose an inter-class affinity contrastive method that captures semantic commonalities among related activities. 
Specifically, we introduce \textit{affinity similarity}, which measures the structural relationships between classes in the embedding space.
Accordingly, these classes that consistently share similar motion patterns are grouped into higher-level superclasses, dubbed Motion Family.
We then propose an inter-class affinity contrastive loss that promotes targeted refinement for semantically related classes.
In addition, a dynamic family-aware temperature schedule is designed to adaptively adjust the penalty strength based on superclass size, thus enhancing representational quality.

Secondly, we propose an intra-class marginal contrastive strategy to mitigate the inherent impact of anomalous positive samples.
This strategy aims to increase the marginal distance constraint between hard positives and their closest negatives, thereby encouraging affinitive aggregation for hard positives. 
Through controlling the minimal margin, an intra-class affinity contrastive loss is designed to achieve a better separation between positives and negatives.
Fig.~\ref{fig:figure1} illustrates the conceptual diagram for our affinity contrastive learning framework.
Extensive experiments are conducted on six popular benchmarks, including NTU RGB+D 60, NTU RGB+D 120, Kinetics-Skeleton, PKU-MMD, and FineGYM for action recognition, as well as CASIA-B for gait recognition and person re-identification.
ACLNet consistently achieves state-of-the-art performance across these benchmark datasets.

Our contributions are summarized as follows:
\begin{itemize}
    \item We introduce ACLNet, a novel affinity contrastive learning network that enhances discriminative representations for skeleton-based human activity understanding.

    \item We propose an inter-class affinity contrastive method that employs the developed affinity metric to capture semantic associations between related activities, thereby enabling globally targeted refinement for hard classes. 

    \item We present an intra-class marginal contrastive strategy to increase the minimal margin between hard positives and negatives, achieving better separation for hard samples.
    
	\item Extensive experiments demonstrate that ACLNet improves upon the current contrastive paradigm by a significant margin, proving to be superior to state-of-the-art methods on six widely used benchmarks.  
\end{itemize}

\section{Related Work}

\subsection{Skeleton-Based Action Recognition}

Early pioneering methods for skeleton-based action recognition primarily focus on recurrent neural networks (RNNs) and convolutional neural networks (CNNs) to predict class labels~\cite{song2018spatio,li2017skeleton}, but they overlook the inherent correlations between joints.
Driven by the structural characteristics of skeleton data, Graph Convolutional Networks (GCNs) have emerged as a leading solution for skeleton-based action recognition~\cite{sun2022human,tu2022joint,liu2025balanced,wei2025va}.
The first notable attempt to introduce the predefined spatial-temporal graph for modeling skeleton data is ST-GCN~\cite{yan2018spatial}.
Although it establishes a strong baseline, the predefined graph topology poses challenges when modeling relationships between joints that lack direct connections, thereby limiting expressive capacity.

Various follow-up studies have explored more flexible modeling of joint relations, including adaptive graphs~\cite{shi2019two,cai2023ske2grid,zhang2024modular,xie2024dynamic}, multi-scale graphs~\cite{liu2020disentangling,yang2025expressive}, and channel-wise graphs~\cite{chen2021channel,chi2022infogcn,duan2022dg,gunasekara2024asynchronous}. 
For instance, 2s-AGCN~\cite{shi2019two} proposes an adaptive graph convolutional network that learns the topology in an end-to-end manner.
Subsequently, InfoGCN~\cite{chi2022infogcn} leverages an information-theoretic objective and multiple modalities to better represent latent information.
Recently, BlockGCN~\cite{zhou2024blockgcn} introduces novel topological encoding schemes, which include static and dynamic encoding to identify and restore the overlooked topologies.

Nevertheless, capturing discriminative semantic features for similar classes remains a challenge for existing methods.
To address this problem, we propose the affinity contrastive learning network, which imposes additional affinity constraints on hard classes and samples, enabling the model to learn more distinctive skeleton representations.

\begin{figure*}[t]
\centering
\includegraphics[width=\linewidth]{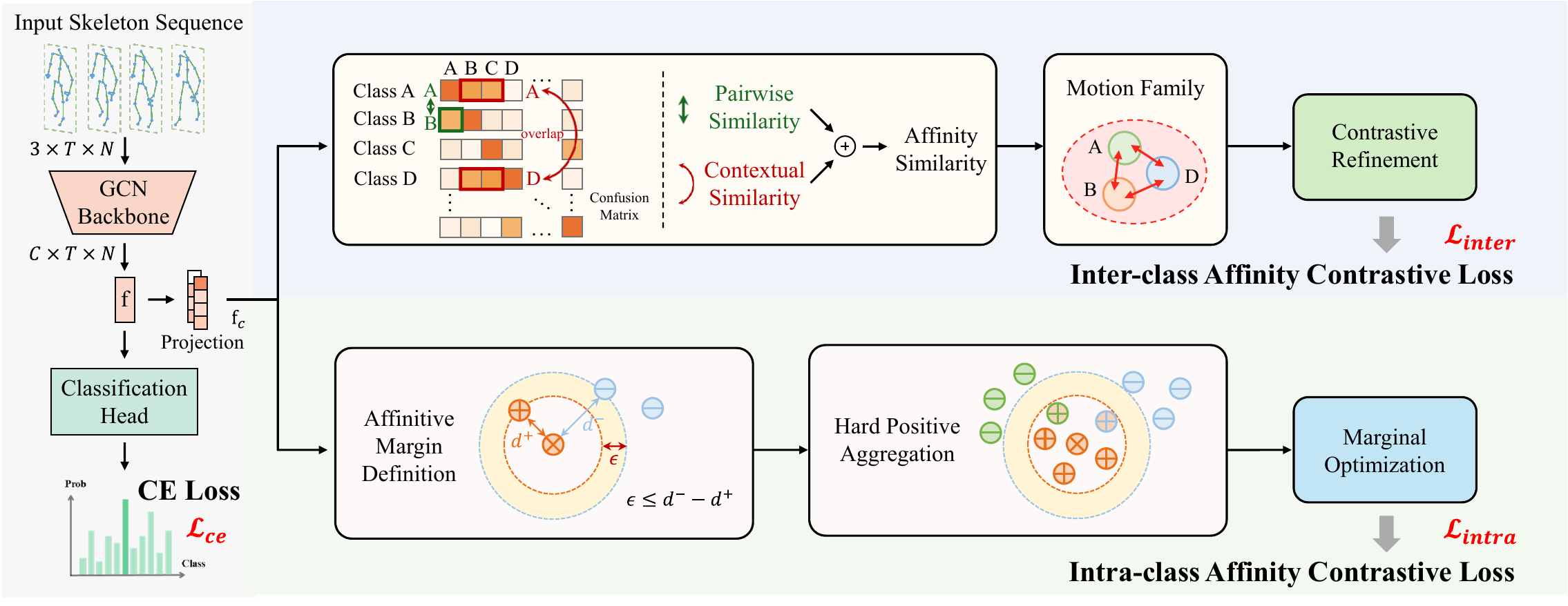}
\caption{ 
The framework of the proposed ACLNet.
The input skeleton sequence is fed into the GCN backbone to extract skeleton feature $f$, which is embedded into a vector by projection for affinity contrastive learning.
Specifically, we introduce affinity similarity to measure the semantic associations between related activities while considering their pairwise and contextual similarities. 
The Motion Family is then constructed to enable targeted refinement for hard classes.
Additionally, we define the affinitive margin to provide accurate control of the minimal distance between the positive sample and the closest negative sample.
By increasing the margin, the optimization strategy helps to improve the separation between hard positives and negatives.
Finally, the two affinity contrastive losses contribute to the construction of a discriminative feature space, effectively improving the accuracy of the model.
}
\label{fig:figure2}
\end{figure*}

\subsection{Skeleton-Based Behavioral Identification}

Skeleton-based behavioral identification has emerged as a key subfield in biometrics, with efforts centered on gait recognition and person re-identification. 
Unlike generic action recognition, this task requires capturing individualized motion patterns that are unique and consistent across instances.
In gait recognition, early skeleton-based approaches like PoseGait~\cite{liao2020model} use 3D keypoints to achieve view invariance. 
Following that, GaitGraph~\cite{teepe2021gaitgraph} and GaitGraph2~\cite{teepe2022towards} adopt multi-branch graph frameworks to model body relationships.
Recent methods such as MSGG~\cite{peng2024learning}, Gait-D~\cite{gao2022gait}, and CycleGait~\cite{li2022strong} leverage GCN-based architectures to achieve considerable performance gains.
In parallel, skeleton-based person re-identification aims to match and retrieve individuals using skeletal representations. 
Early works manually design skeleton descriptors based on anthropometric and gait attributes~\cite{liu2015enhancing}.
Rao et al.~\cite{rao2020self} utilize an LSTM-based encoder with attention mechanisms (AGE) to learn skeletal features. 
Its extension SGELA~\cite{rao2021self} integrates skeletal pretext tasks and inter-sequence contrastive learning to enhance representations.
Recently, TranSG~\cite{rao2023transg} proposes a Transformer-based skeleton graph contrastive learning framework to capture complex skeletal relations.
Compared with these methods, we introduce the affinity contrastive learning to highlight subtle differences between classes, achieving effective biometric identification.
The proposed affinity modeling paradigm opens new avenues for fine-grained activity understanding, with potential applications in various behavioral biometrics tasks.

\subsection{Contrastive Learning}

Contrastive learning is a classical self-supervised representation method that has been shown to improve both performance and robustness of downstream tasks in diverse fields~\cite{he2020momentum,chen2020simple,song2021spatio,khaertdinov2022dynamic}.
Methods such as MoCo~\cite{he2020momentum} and SimCLR~\cite{chen2020simple} propose ingenious paradigms to focus on semantic representations through this contrastive learning manner.
Another line of work~\cite{kim2022self,barbano2022unbiased} focuses on metric learning, consistently bootstrapping the representation by discovering the hard positive and negative samples.
Inspired by these contrastive paradigms, our work leverages the developed affinity metric and the affinitive marginal constraint to encourage the model to learn discriminative information.
In the field of skeleton-based human activity understanding, there are two research perspectives on the application of contrastive learning.
In self-supervised~\cite{zhang2023prompted,gunasekara2025spatio} and unsupervised settings~\cite{lin2023actionlet}, prior approaches utilize skeleton transformations to construct diverse positive and negative pairs, aiming to maintain consistency within the embedding space.
In parallel, contrastive learning has also demonstrated considerable potential in fully supervised scenarios~\cite{huang2023graph,zhou2023learning,chang2024wavelet,bian2024class}.
Regarding the distinction of similar actions, the methodology of current works can be summarized as finding differences in actions from a specific perspective.
For instance, Huang et al.~\cite{huang2023graph} explicitly explore the global context information across all sequences, and Zhou et al.~\cite{zhou2023learning} propose spatio-temporal feature refinement to improve discriminative representations.

In contrast, our objective is to inform the recognition model which actions are semantically confusing, thereby prompting the model to focus on distinguishing them and spontaneously finding differences.
The affinity information contains rich supervisory signals that can provide significant disambiguation clues.
Consequently, this enables the model to effectively capture the distinctive semantics between activities.

\section{Method}

In this section, we begin by briefly introducing the preliminary concepts of skeleton-based human activity understanding using GCNs. 
We then provide a detailed description of the proposed affinity contrastive learning framework. 
An overview of the proposed ACLNet is depicted in Fig.~\ref{fig:figure2}.

\subsection{Preliminaries}
\label{method_1}

The input for skeleton-based approaches is a sequence of skeletons spanning $T$ frames, where each skeleton consists of $N$ body joints.
The human skeleton is typically represented as a graph $\mathcal{G}=(\mathcal{V},\mathcal{E})$, where $\mathcal{V}=\left\{v_{1},v_{2}, \dots,v_{N}\right\} $ is the set of body joints, $\mathcal{E}$ represents the connectivity among these joints. 
In practice, $\mathcal{E}$ is implemented as the adjacency matrix $\mathbf{A}\in\mathbb{R}^{N\times N}$, where each element $a_{ij}$ reflects the strength of the correlation between joints $v_i$ and $v_j$.
The framework of GCN models can be perceived as a backbone network consisting of $L$ stacked graph convolutional layers, followed by a classification head.
The extracted semantic feature of joint motions can be denoted as $\mathbf{X} \in \mathbb{R}^{C\times T\times N}$, where $C$ is the dimension of motion features.

Mathematically, the translation function of the $l$-th layer within the backbone network can be written as,
\begin{equation}
  \mathbf{X}^{(l)} = \sigma (\mathbf{A}^{(l)}\mathbf{X}^{(l-1)}\Theta^{(l)})
  \label{eq_1}
\end{equation}
where $\mathbf{A}\in\mathbb{R}^{N\times N}$ is the adjacency matrix representing the correlations between $N$ joints, $\Theta^{(l)} \in \mathbb{R}^{C^{(l-1)}\times C^{(l)}}$ denotes the learnable weight of the convolutional operation, and $\sigma$ indicates the ReLU activation function.
After obtaining the final hidden representation $\mathbf{X}^{(L)}$, a classification network is employed to determine the prediction label $\hat{\mathbf{y}}\in \mathbb{R}^c$.
Here $c$ denotes the total number of classes.
The cross-entropy loss $\mathcal{L}_{ce}$ is then applied to supervise the predicted class
\begin{equation}
  \mathcal{L}_{ce} = -\sum_{i}^{c} \mathbf{y_{i}} \log \hat{\mathbf{y_{i}}} 
  \label{eq_1_2}
\end{equation}
where $\mathbf{y_{i}}$ is the one-hot ground-truth label.
Additionally, a projection layer is employed to embed the last hidden representation into the vector $\mathbf{f}_{c}$ within the contrastive feature space, which is utilized in the computation of objective functions for the proposed affinity contrastive learning network.

\subsection{Inter-class Affinity Contrastive Learning}
\label{method_2}

To tackle the challenge of differentiating between activities that are prone to misclassification, we investigate the intricate inter-class relationships and introduce the concept of Motion Family.
Specifically, Motion Family is the superclass formed by clustering based on the proposed \textit{affinity similarity}, which represents the set of associated activities with structural commonalities.
In the following subsections, we provide a detailed explanation of \textit{affinity similarity} and then introduce the inter-class affinity contrastive loss.

\subsubsection{Affinity Similarity Definition}

The quality of the motion semantic associations is vital since they determine the clustering of the Motion Family.
However, computing reliable semantic associations is non-trivial, especially at the early stages of the training process.
To overcome this issue, we propose \textit{affinity similarity} to estimate high-quality semantic associations between related activities.

Our key idea is to leverage not only the direct pairwise relations between the two activity classes, but also the indirect semantic commonalities via their overlap.
Intuitively, if two activities share many similar classes, they should be considered to have a hidden commonality and thus be within the same superclass space.
During training, recognition models can implicitly create a similarity graph between activities, and our method introduces affinity constraints on top of this graph, resulting in self-taught clustering relationships. 
Such implicit inter-class affinity helps to capture structural commonalities between classes and provides rich supervisory signals for further contrastive learning efficiently and concisely.

In practice, we begin by gathering misclassified samples to construct the initial confusion matrix between activity classes, which discovers the direct pairwise associations.
We then capture indirect hidden commonalities by considering their overlaps in the confusion matrix.
The proposed \textit{affinity similarity} is defined as the combination of direct pairwise similarity and indirect contextual similarity.
Accordingly, we split the calculation into two sequential steps.

\subsubsection{Affinity Similarity Calculation}

The first step involves calculating pairwise associations between classes by identifying confusing classes.
A global statistics matrix $H$, initialized to zero, represents the pairwise relations between activities. 
Here $H_{i,j}$ indicates the number of samples belonging to class $i$ that are misclassified as class $j$.
Given all the inputs, this matrix helps to establish the global pairwise associations.

To reduce the impact of predictive randomness, we focus on the top $K$ most similar classes, which are highly correlated, as pairwise candidates.
The pairwise association set for class $i$ is further defined as:
\begin{equation}
	\mathcal{N}_{K}(i) = \{j|j \in {\rm{Top}}_K(H_{i, \cdot}), j \neq i \}
	\label{eq_2_temp}
\end{equation}
where $H_{i, \cdot}$ indicates the statistics of class $i$ with all other classes, and $\mathcal{N}_{K}(i)$ represents $K$ classes with higher values in the statistics ${\rm{Top}}_K(H_{i, \cdot})$.
Statistically, $K$ is set to 10.
Then we establish the binary matrix $\mathbb{I}_{\mathcal{N}_{K}}(i,j)$ indicating the pairwise similarity between class $i$ and class $j$, which could be expressed as:
\begin{equation}
	\mathbb{I}_{\mathcal{N}_{K}}(i,j) = \left\{
		\begin{array}{ll}
		1, & {\rm{if}} \, j \in \mathcal{N}_{K}(i) \\
		0, & {\rm{if}} \, j \notin \mathcal{N}_{K}(i) 
		\end{array}
		\right.
	\label{eq_3}
\end{equation}

Since the network is not yet mature enough to grasp the semantic relations between classes at the early stages of training, the preliminary associations are often compromised.
In addition, the potential structural commonalities could provide implicit supervisory information.
Therefore, we propose calculating the \textit{affinity similarity}, as detailed below.

The second step aims to explore the indirect semantic associations and capture the final affinity relationship.
We refine the contextual information by counting the number of neighbor activities that the two classes have in common. 
If two classes share many similar activities, they are more likely to have structural commonalities.

Consisting of pairwise and contextual similarity, the \textit{affinity similarity} between classes $i$ and class $j$ could be defined as:
\begin{equation}
	\begin{split}
	& w_{ij} = \frac{\mathbb{I}_{\mathcal{N}_{K}}(i,j)}{2} + \frac{M(i,j)}{\sum_{p} \mathbb{I}_{\mathcal{N}_{K}}(i,p)} \, , \\
	& {\rm where} \, \, M(i,j) = \sum_{p = 1}^{K} \mathbb{I}_{\mathcal{N}_{K}}(i,p)\mathbb{I}_{\mathcal{N}_{K}}(j,p) \, .
	\end{split}
	\label{eq_4}
\end{equation}

Here $p$ denotes the index of pairwise activity for class $i$, and $M(i,j)$ means the total number of classes that $i$ and $j$ have in common through the AND operation.
$\sum_{p} \mathbb{I}_{\mathcal{N}_{K}}(i,p)$ represents the size of $\mathbb{I}_{\mathcal{N}_{K}}(i,j)$, generally $\sum_{p} \mathbb{I}_{\mathcal{N}_{K}}(i,p)=K$.
In practice, we accordingly halve the pairwise correlation weight and then add it to the overlap value, finally obtaining the \textit{affinity similarity}.

\subsubsection{Motion Family Refinement}

So far, the direct and indirect correlations between classes have been effectively explored. 
We then construct the Motion Family $W(i)$ as follows:
\begin{equation}
	W(i) = \{j | w_{ij} > \frac{n_{a}}{K}  \}
	\label{eq_5}
\end{equation}
where $n_{a}$ denotes the overlap threshold.

Thereafter, we refine the distinctive representations across the Motion Family, which represents the set of associated activities with common properties.
For these hard classes with semantic consistency, the targeted refinement is carried out to better capture the feature differences between the member classes in the Motion Family.
For each member class, we define the class representation $\mathbf{m_i}$, and update it with Exponential Moving Average~(EMA):
\begin{equation}
	\mathbf{m_i} = \gamma \cdot \mathbf{m_i} + (1-\gamma) \cdot \frac{1}{n_k} \sum_{k = 1}^{n_k} f_k^{i}
	\label{eq_6}
\end{equation}
where $f_k^{i}\in\mathbb{R}^{d}$ represents the $k$-th feature of class $i$ within the input batch, $\gamma$ is the momentum term, and $n_k$ is the total number of samples. 
Ideally, newly arrived samples of class $i$ should converge with $\mathbf{m_i}$ and differ from the representations of other classes.
Along with the process, $\mathbf{m_i}$ gradually becomes a stable estimation of the clustering center for class $i$, establishing the foundation for the subsequent refinement.

Lastly, we propose the inter-class affinity contrastive loss to optimize the inter-class learning objective.
Let $f_{\mu}^{i}$ denote the feature of sample $\mu$ belonging to class $i$.
The inter-class affinity contrastive loss could be formulated as:
\begin{equation}
	\mathcal{L}_{inter} = - \log \frac{ \exp (f_{\mu}^{i} \cdot \mathbf{m_i} / {\tau}_w )} {\exp (f_{\mu}^{i} \cdot \mathbf{m_i} / {\tau}_w ) + \sum\limits_{a \in W(i)} \exp (f_{\mu}^{i} \cdot \mathbf{m_a} / {\tau}_w )}
	\label{eq_7}
\end{equation}
where $a$ is the member class index, the $\cdot$ symbol denotes the inner product operation, and $\mathbf{m_i}, \mathbf{m_a}$ denote the corresponding class representations.

\subsubsection{Discussion}

The proposed Motion Family is more of a conceptual framework than a direct mechanism for physically bringing similar classes closer together. 
Here, the affinity relationships serve as effective supervisory signals that guide the model to identify semantically related classes and focus on their refinement.
Meanwhile, to address the inherent variability within each class, we employ average aggregation and cross-batch momentum updates, which effectively mitigate the impact of intra-class diversity and ensure stable updates.

\subsection{Family-Aware Temperature Schedule}
\label{method_3}

In contrastive learning, the models are trained to ensure that embeddings of different classes are repelled while embeddings of the same class are attracted.
The strength of these attractive and repelling forces between samples is controlled by the temperature parameter, which has been found to crucially impact the quality of the learned representations~\cite{wang2020understanding}.
Motivated by this, we consider that the penalty strength associated with different sizes of the Motion Family should also be adaptive.
Therefore, we employ a dynamic temperature ${\tau}_w$ during training.
Through the dynamic schedule, the representation quality could be improved without additional cost.

In practice, we modify ${\tau}_w$ according to the simplistic interval incremental schedule determined by the actual superclass size $N_w$.
The hyperparameter $K$ is set to 10, which corresponds to the threshold for the superclass size.
The dynamic temperature ${\tau}_w$ could be formulated as:
\begin{equation}
	{\tau}_w = \left\{
		\begin{array}{ll}
		0.1, 	& {\rm{if}} \, \, N_w \leq K\\
		0.5, 	& {\rm{if}} \, \, K < N_w \leq 2K \\
		1.0, 	& {\rm{if}} \, \, N_w > 2K 
		\end{array}
		\right.
	\label{eq:schedule}
\end{equation}

Specifically, a relatively larger $\tau$ (${\tau}_w=1.0$) could increase the margin between clusters and facilitate the cluster-wise discrimination.
In contrast, when the superclass sizes become small, a smaller $\tau$ (${\tau}_w=0.1$) could be used to amplify differences in similarity for hard negative samples in the embedding space, contributing to the instance-specific refinement within the superclass.
For better smooth optimization, we add the ${\tau}_w=0.5$ setting.
Such a schedule results in a constant `family-aware switching’ between an emphasis on different superclasses, thereby ensuring that the model consistently improves separation between member classes.

\subsection{Intra-class Affinity Contrastive Learning}
\label{method_4}

Building upon the inter-class constraints, we further improve the intra-class representations.
In general, abundant sample diversity inevitably introduces intrinsic noise, which will result in the presence of hard positives that themselves are easily confused with other classes.
These hard positives, along with negatives from similar classes, would lead to the accumulation of errors and degrade the overall performance.

To this end, we present an intra-class margin-based learning objective that provides more accurate control of the minimal margin between the positive sample and the closest negative sample.
As shown in Fig.~\ref{fig:figure2}, the proposed marginal strategy can be regarded as an affinitive aggregation for all positives with the class, thereby achieving a better separation between hard positives and negatives.

Let $x$ be the anchor of the original skeleton sample, $x_u^+$ a positive sample, $x_v^-$ a negative sample, and $N_v$ the number of negative samples.
Here, positive and negative samples refer to samples belonging to the same and different classes.
$s(f_m,f_n)$ is defined as the cosine similarity between the samples $m$ and $n$.
Since ${\| f_m \|}_{2} = {\| f_n \|}_{2} = 1$, the L2-distance $d(f_m,f_n)={\| f_m-f_n \|}_{2}^{2}$ is smaller, and cosine similarity is larger.
We denote $d(f_x,f_{x^+})$ as $d^+$, $d(f_x,f_{x_v^-})$ as $d_v^-$, and correspondingly, $s(f_x,f_{x^+})$ as $s^+$, $s(f_x,f_{x_v^-})$ as $s_v^-$.
Consider first the case of a single positive sample $x^+$. 
For the intra-class divergence, the following condition needs to be satisfied:
\begin{equation}
	d_v^- - d^+ \geq \epsilon \quad \forall v
	\label{eq_8}
\end{equation}
where $\epsilon>0$ is the margin between positive and negative samples.
The condition could be further derived:
\begin{equation}
	d^+ - d_v^- \leq -\epsilon \Longleftrightarrow  s_v^- - s^+ \leq -\epsilon \quad \forall v
	\label{eq_9}
\end{equation}

Based on InfoNCE~\cite{oord2018representation}, the above constraint could be transformed into an optimization problem with the $\max$ operator and the smooth approximation $LogSumExp$ (LSE) operator:
\begin{equation}
	\begin{split}
	& \max (-\epsilon, {\{ s_v^- - s^+ \}}_{v=1,...,N_v}) \\
	& \approx - \log (\frac{\exp (s^+)}{\exp (s^+ - \epsilon) + \sum_{v} \exp (s_v^-) })
	\end{split}
	\label{eq_10}
\end{equation}

Then we generalize the above optimization to multiple positive samples to derive the following constraints:
\begin{equation}
	s_v^- - s_u^+ \leq -\epsilon \quad \forall u,v
	\label{eq_11}
\end{equation}
\begin{equation}
	\sum\limits_{u} \max (-\epsilon, {\{ s_v^- - s_u^+ \}}_{v=1,...,N_v})
	\label{eq_12}
\end{equation}

Therefore, the proposed intra-class marginal contrastive loss could be formulated as:
\begin{equation}
	\mathcal{L}_{intra} = - \sum\limits_{u} \log (\frac{\exp (s_u^+ / \tau)}{\exp ((s_u^+ / \tau) - \epsilon) + \sum_{v} \exp (s_v^- / \tau) })
	\label{eq_13}
\end{equation}
where $\epsilon$ applies to all intra-class positives, and could achieve the separation between hard positive and negative samples.

\subsection{Overall Objective Functions}
\label{method_5}

Finally, the overall loss function used to train the model could be written as: 
\begin{equation}
	\mathcal{L} = \mathcal{L}_{ce} + \lambda_{1} \mathcal{L}_{inter} + \lambda_{2} \mathcal{L}_{intra} 
	\label{eq_14}
\end{equation}
where $\mathcal{L}_{ce}$ is the cross-entropy loss used to supervise the predicted class.
For balance, $\lambda_{1}$ and $\lambda_{2}$ are the weights assigned to the inter-class affinity contrastive loss $\mathcal{L}_{inter}$ and the intra-class marginal contrastive loss $\mathcal{L}_{intra}$.

\section{Experiments}

\subsection{Datasets}
\label{experiments_1}

\noindent
\textbf{NTU RGB+D 60}~\cite{shahroudy2016ntu} contains 56,880 indoor captured skeleton action samples, performed by 40 different subjects and classified into 60 classes. 
This dataset recommends two evaluation protocols: (1) cross-subject (X-Sub): train data are performed by 20 subjects, and test data are performed by the other 20 subjects. (2) cross-view (X-View): train data from camera views 2 and 3, and test data from camera view 1.

\noindent
\textbf{NTU RGB+D 120}~\cite{liu2019ntu} contains 114,480 skeleton action samples over 120 classes
There are also two protocols: (1) cross-subject (X-sub): skeleton samples from 53 subjects are used for training, while the remaining 53 are used for testing. (2) cross-setup (X-set): training data comes from 16 even setup IDs, and testing data comes from 16 odd setup IDs.

\noindent
\textbf{Kinetics-Skeleton} is derived from the Kinetics 400 video dataset~\cite{kay2017kinetics}, utilizing the pose estimation toolbox to extract 240,436 training and 19,796 evaluation skeleton samples across 400 classes.
We use the skeleton data released by Duan~et~al.~\cite{duan2022pyskl} to evaluate the model.
Following the standard evaluation protocol, Top-1 and Top-5 accuracies are reported.

\noindent
\textbf{PKU-MMD}~\cite{liu2017pku} is a comprehensive dataset for human action recognition, comprising more than 20,000 samples across 51 categories. 
For the X-Sub setting, 57 subjects are designated for training, while 9 subjects are reserved for testing.
For the X-View setting, the middle and right views are used for training, with the left view serving as the test set.

\noindent
\textbf{FineGYM}~\cite{shao2020finegym} is a large-scale fine-grained action recognition dataset with 29,000 videos of 99 gymnastic action classes, which requires action recognition methods to distinguish different sub-actions within the same video.
We use the skeleton data provided by Duan~et~al.~\cite{duan2022pyskl}.
The mean class Top-1 accuracy is reported in the evaluation protocol.

\noindent
\textbf{CASIA-B}~\cite{yu2006framework} is a multi-view human gait dataset comprising 124 subjects, each captured from 11 camera views. 
For each angle, every subject has 10 sequences under three walking conditions: 6 normal walking (NM), 2 with a bag (BG), and 2 with clothes (CL).
In addition, we adopt the commonly-used probe and gallery settings~\cite{liu2015enhancing} for person re-identification.
The settings are Normal (N-N), Bags (B-B), Clothes (C-C).
In cross-condition settings, (C-N) denotes using the Clothes probe set and Normal gallery set, and (B-N) denotes matching the Bags probe set with the Normal gallery set.
The Rank-1 accuracy is used to evaluate model performance.

\begin{table}[t]
\centering
\caption{ 
Performance comparisons against the state-of-the-art methods on the NTU RGB+D 60 dataset in terms of classification accuracy (\%). 
}
\setlength\tabcolsep{8pt}
\renewcommand\arraystretch{1.1}
\resizebox{\linewidth}{!}{
\begin{tabular}{l r c c}
	\toprule
	\multirow{2}{*}{Methods} & \multirow{2}{*}{Publication} & \multicolumn{2}{c}{NTU RGB+D 60} \cr
	& & X-Sub & X-View \cr
	\midrule
	ST-GCN \cite{yan2018spatial} & AAAI 2018 & 81.5 &  88.3 \\
	2s-AGCN \cite{shi2019two} & CVPR 2019 & 88.5 & 95.1 \\
	MS-G3D \cite{liu2020disentangling} & CVPR 2020 & 91.5 & 96.2 \\
	CTR-GCN \cite{chen2021channel} & ICCV 2021 & 92.4 & 96.8 \\
	EfficientGCN-B4 \cite{song2022constructing}  & TPAMI 2022 & 92.1 & 96.1 \\
	InfoGCN \cite{chi2022infogcn} & CVPR 2022 & 93.0 & 97.1 \\
	SkeletonGCL \cite{huang2023graph} & ICLR 2023 & 92.8 & 97.1 \\
	FR-Head \cite{zhou2023learning} & CVPR 2023 & 92.8 & 96.8 \\
	HD-GCN \cite{lee2023hierarchically} & ICCV 2023 & 93.4 & 97.2 \\
	DS-GCN \cite{xie2024dynamic} & AAAI 2024 & 93.1 & 97.5 \\
	BlockGCN \cite{zhou2024blockgcn} & CVPR 2024 & 93.1 & 97.0 \\
	VA-AR \cite{wei2025va} & AAAI 2025 & 93.1 & 97.2 \\
	\midrule
	Ours & & \textbf{93.6} & \textbf{97.7} \\
	\bottomrule
\end{tabular}
}
\label{tab:ntu_60}
\end{table}

\begin{table}[ht]
\centering
\caption{ 
Performance comparisons against the state-of-the-art methods on the NTU RGB+D 120 dataset in terms of classification accuracy (\%). 
}
\setlength\tabcolsep{8pt}
\renewcommand\arraystretch{1.1}
\resizebox{\linewidth}{!}{
\begin{tabular}{l r c c}
	\toprule
	\multirow{2}{*}{Methods} & \multirow{2}{*}{Publication} & \multicolumn{2}{c}{NTU RGB+D 120} \cr
	& & \,X-Sub & X-Set \cr
	\midrule
	ST-GCN \cite{yan2018spatial} & AAAI 2018 & 70.7 & 73.2 \\
	2s-AGCN \cite{shi2019two} & CVPR 2019 & 82.5 & 84.2 \\
	MS-G3D \cite{liu2020disentangling} & CVPR 2020 & 86.9 & 88.4 \\
	CTR-GCN \cite{chen2021channel} & ICCV 2021 & 88.9 & 90.6 \\
	EfficientGCN-B4 \cite{song2022constructing}  & TPAMI 2022 & 88.7 & 88.9 \\
	InfoGCN \cite{chi2022infogcn} & CVPR 2022 & 89.8 & 91.2 \\
	SkeletonGCL \cite{huang2023graph} & ICLR 2023 & 89.8 & 91.2 \\
	FR-Head \cite{zhou2023learning} & CVPR 2023 & 89.5 & 90.9 \\
	HD-GCN \cite{lee2023hierarchically} & ICCV 2023 & 90.1 & 91.6 \\
	DS-GCN \cite{xie2024dynamic} & AAAI 2024 & 89.2 & 91.1 \\
	BlockGCN \cite{zhou2024blockgcn} & CVPR 2024 & 90.3 & 91.5 \\
	VA-AR \cite{wei2025va} & AAAI 2025 & 90.3 & 91.5 \\
	\midrule
	Ours & & \textbf{90.7} & \textbf{92.3} \\
	\bottomrule
\end{tabular}
}
\label{tab:ntu_120}
\end{table}

\subsection{Implementation Details}
\label{experiments_2}

We use PyTorch and $1\times \,$RTX 3090 GPU for experiments.
We take the contrastive learning model FR-Head~\cite{zhou2023learning} as the baseline. 
The SGD optimizer is employed with a Nesterov momentum of 0.9 and a weight decay of $5\times10^{-4}$. 
The epoch number is 150, and we set the batch size to 64.
The initial learning rate is set to 0.1 with a cosine learning rate scheduler.
To avoid instability in early training, we perform affinity calculations beginning from epoch 30.
The computational complexity is closely tied to the batch size and the number of classes.
In addition, $\gamma$ in Eq.~\ref{eq_6} is set to $0.9$, and $\tau$ in Eq.~\ref{eq_13} is set by default to $0.1$. 
The dimension of feature vectors $\mathbf{f}_{c}$ for contrastive learning is set to 256, and the weights $\lambda_1$ and $\lambda_2$ in Eq.~\ref{eq_14} are set to $0.1$.
We use the framework and data pre-processing procedures outlined by Duan~et~al.~\cite{duan2022pyskl}, which perform efficient spatial and temporal augmentations.
We follow the gait recognition settings and data augmentations from GaitGraph~\cite{teepe2021gaitgraph}.
For person re-identification, we follow TranSG~\cite{rao2023transg} and set the sequence length to 40 frames.
The random seed is fixed to ensure experiment reproducibility.

\begin{table}[t!]
\centering
\caption{ 
Performance comparisons against the state-of-the-art methods on the Kinetics-Skeleton dataset in terms of classification accuracy (\%). 
}
\setlength\tabcolsep{10.8pt}
\renewcommand\arraystretch{1.1}
\resizebox{\linewidth}{!}{
\begin{tabular}{l r c c}
	\toprule
	\multirow{2}{*}{Methods} & \multirow{2}{*}{Publication} & \multicolumn{2}{c}{Kinetics-Skeleton} \cr
	& & Top-1 & \,Top-5 \cr
	\midrule
	ST-GCN \cite{yan2018spatial} & AAAI 2018 & 30.7 & 52.8 \\
	2s-AGCN \cite{shi2019two} & CVPR 2019 & 36.1 & 58.7 \\
	MS-G3D \cite{liu2020disentangling} & CVPR 2020 & 38.0 & 60.9 \\
	UPS \cite{foo2023unified} & CVPR 2023 & 40.5 & 63.3 \\
	STFD-Net \cite{zhao2024glimpse} & TCSVT 2024 & 48.1 & - \\
	DS-GCN \cite{xie2024dynamic} & AAAI 2024 & 50.6 & - \\
	\midrule
	Ours & & \textbf{52.1} & \textbf{75.9} \\
	\bottomrule
\end{tabular}
}
\label{tab:k400}
\end{table}

\begin{table}[t!]
\centering
\caption{
Performance comparisons with the state-of-the-art methods on PKU-MMD in terms of classification accuracy (\%). 
}
\setlength\tabcolsep{11pt}
\renewcommand\arraystretch{1.1}
\resizebox{\linewidth}{!}{
\begin{tabular}{l r c c}
\toprule
\multirow{2}{*}{Methods} & \multirow{2}{*}{Publication} & \multicolumn{2}{c}{PKU-MMD} \cr
& & X-Sub & X-View \cr
\midrule
STA-LSTM \cite{song2018spatio} & TIP 2018 & 86.9 & 92.6 \\
RF-Action \cite{li2019making} & ICCV 2019 & 92.9 & 94.4 \\
ActCLR \cite{lin2023actionlet} & CVPR 2023 & 90.0 & 91.6 \\
DMMG \cite{guan2023dmmg} & TIP 2023 & 91.7 & - \\
HA-GNN \cite{geng2024hierarchical} & TMM 2024 & 92.2 & - \\
AJTP \cite{gunasekara2024asynchronous} & TCSVT 2024 & 95.0 & 96.7 \\
\midrule
Ours &  & \textbf{97.3} & \textbf{98.7} \\
\bottomrule
\end{tabular}
}
\label{tab:PKU-MMD}
\end{table}

\begin{table}[t!]
\centering
\caption{
Performance comparisons with the state-of-the-art methods on FineGYM in terms of classification accuracy (\%).
}
\setlength\tabcolsep{14pt}
\renewcommand\arraystretch{1.1}
\resizebox{\linewidth}{!}{
\begin{tabular}{l r c}
\toprule
Methods & Publication & FineGYM \cr
\midrule
LT-S3D \cite{xie2018rethinking} & ECCV 2018 & 88.9 \\
PoseConv3D \cite{duan2022revisiting} & CVPR 2022 & 94.3 \\
Ske2Grid \cite{cai2023ske2grid} & ICML 2023 & 91.8 \\
SkeletonMAE \cite{yan2023skeletonmae} & ICCV 2023 & 91.8 \\
VA-AR \cite{wei2025va} & AAAI 2025 & 92.8 \\
\midrule
Ours &  & \textbf{96.0} \\
\bottomrule
\end{tabular}
}
\label{tab:FineGYM}
\end{table}

\begin{table}[ht]
\centering
\caption{
Performance comparisons with the state-of-the-art skeleton-based gait recognition methods on CASIA-B in terms of averaged Rank-1 accuracy (\%).
}
\setlength\tabcolsep{13pt}
\renewcommand\arraystretch{1.1}
\resizebox{\linewidth}{!}{
\begin{tabular}{l c c c c}
\toprule
\multirow{2}{*}{Methods} & \multicolumn{4}{c}{CASIA-B} \cr
& NM & BG & CL & Avg \cr
\midrule
PoseGait \cite{liao2020model} & 68.7 & 44.5 & 36.0 & 49.7 \\
GaitGraph \cite{teepe2021gaitgraph} & 87.7 & 74.8 & 66.3 & 76.3 \\
GaitGraph2 \cite{teepe2022towards} & 82.0 & 73.2 & 63.6 & 72.9 \\
MSGG \cite{peng2024learning} & 93.0 & 78.1 & 68.3 & 79.8 \\
Gait-D \cite{gao2022gait} & 91.6 & 79.0 & 72.0 & 80.9 \\
CycleGait \cite{li2022strong} & 92.8 & 84.0 & 78.7 & 85.2 \\
\midrule
Ours & \textbf{94.7} & \textbf{85.4} & \textbf{85.4} & \textbf{88.5} \\
\bottomrule
\end{tabular}
}
\label{tab:CASIA-B-gait}
\end{table}

\begin{table}[ht]
\centering
\caption{
Performance comparison with the state-of-the-art person re-identification methods on CASIA-B in terms of Rank-1 accuracy (\%).
}
\setlength\tabcolsep{11pt}
\renewcommand\arraystretch{1.1}
\resizebox{\linewidth}{!}{
\begin{tabular}{l c c c c c}
\toprule
\multirow{2}{*}{Methods} & \multicolumn{5}{c}{Probe-Gallery} \cr
& N-N & B-B & C-C & C-N & B-N \cr
\midrule
ELF \cite{gray2008viewpoint} & 12.3 & 5.8 & 19.9 & 5.6 & 17.1 \\
MLR \cite{liu2015enhancing} & 16.3 & 18.9 & 25.4 & 20.3 & 31.8 \\
AGE \cite{rao2020self} & 20.8 & 37.1 & 35.5 & 14.6 & 32.4 \\
SGELA \cite{rao2021self} & 71.8 & 48.1 & 51.2 & 15.9 & 36.4 \\
TranSG \cite{rao2023transg} & 78.5 & 67.1 & 65.6 & 23.0 & 44.1 \\
\midrule
Ours & \textbf{82.8} & \textbf{68.5} & \textbf{68.0} & \textbf{25.2} & \textbf{46.5} \\
\bottomrule
\end{tabular}
}
\label{tab:CASIA-B-reid}
\end{table}

\subsection{Comparison with State-of-the-Art Methods}
\label{experiments_3}

We compare ACLNet with state-of-the-art methods on six benchmark datasets, including NTU RGB+D 60, NTU RGB+D 120, Kinetics-Skeleton, PKU-MMD, FineGYM, and CASIA-B.
The proposed method consistently achieves state-of-the-art performance in all scenarios.

The results on the NTU RGB+D 60 dataset are illustrated in Table~\ref{tab:ntu_60}. 
ACLNet achieves the best classification accuracy of 93.6\% under the X-Sub setting and 97.7\% under the X-View setting.
Furthermore, on the challenging NTU RGB+D 120 dataset, ACLNet demonstrates comparable performance, achieving accuracies of 90.7\% under the X-Sub setting and 92.3\% under the X-Set setting, as shown in Table~\ref{tab:ntu_120}.
For Kinetics-Skeleton, it is evident from Table~\ref{tab:k400} that the proposed method has delivered superior performance.
In addition, according to the results in Table~\ref{tab:PKU-MMD} and Table~\ref{tab:FineGYM}, the significant improvement of the proposed ACLNet on the PKU-MMD and FineGYM datasets demonstrates its powerful ability to model diverse and complex actions.
As presented in Table~\ref{tab:CASIA-B-gait}, our method achieves competitive results compared with the representative skeleton-based gait recognition methods on CASIA-B. 
For person re-identification, the results in Table~\ref{tab:CASIA-B-reid} verify the effectiveness of ACLNet to learn discriminative patterns, and demonstrate the potential for biometric applications.

\begin{table}[t!]
\centering
\caption{
Comparisons of classification accuracy (\%) when applying different components under the NTU-60 X-Sub setting with the joint modality.
}
\renewcommand\arraystretch{1.2}
\resizebox{\linewidth}{!}{
\begin{tabular}{ c c c c c c }
\toprule
Baseline & Inter-ACL & FAT & Intra-ACL & Params & X-Sub \\
\midrule
\checkmark &  &  &  & 3.56M & 90.3 \\
\checkmark & \checkmark &  && 3.71M & 90.9 ($\uparrow$ 0.6) \\
\checkmark & \checkmark & \checkmark &  & 3.71M & 91.1 ($\uparrow$ 0.8) \\
\checkmark &  &  & \checkmark & 3.68M & 90.8 ($\uparrow$ 0.5) \\
\checkmark & \checkmark & \checkmark & \checkmark& 3.86M & \textbf{91.4 ($\uparrow$ 1.1)} \\
\bottomrule
\end{tabular}
}
\label{tab:ablation_1}
\end{table}

\subsection{Ablation Study}
\label{experiments_4}

In this part, we conduct ablation studies to evaluate the effectiveness of the proposed method.
The experiments are performed on the NTU RGB+D 60 with the joint modality under the X-Sub and X-View settings.

\begin{figure*}[t!]
\centering
\includegraphics[width=\linewidth]{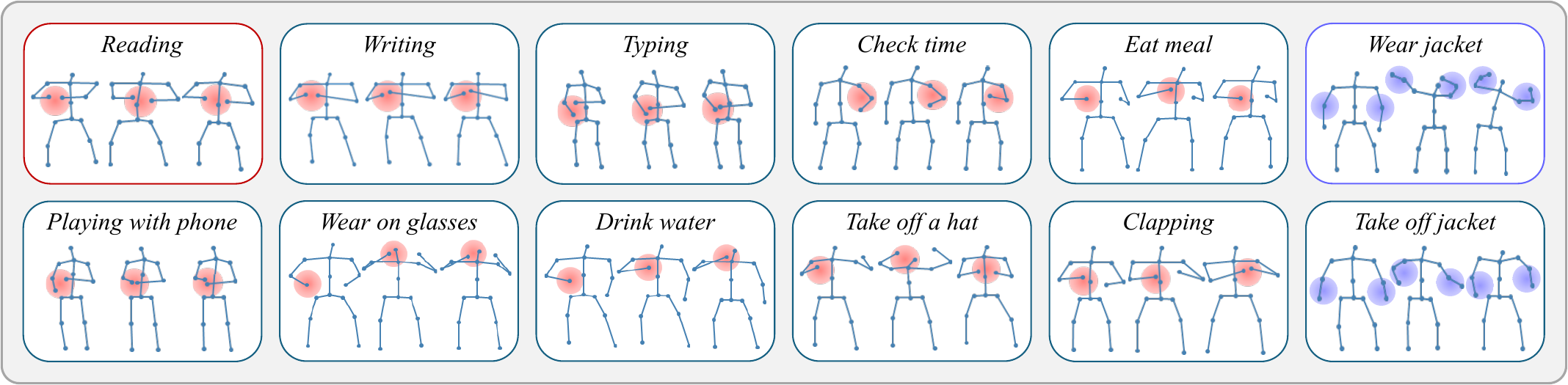}
\caption{
Examples of Motion Family corresponding to the anchor actions `reading' and `wear jacket'. 
Red and blue areas reflect the notable body parts with high learned weights, which indicate the structural commonality links (e.g., hand-related and arm-related) in the skeleton sequences.
}
\vspace{-0.5mm}
\label{fig:figure3}
\end{figure*}

\begin{figure*}[t!]
\centering
\subfloat[Accuracy Vs. Threshold ($n_{a}$)]
{\includegraphics[width=.24\linewidth]{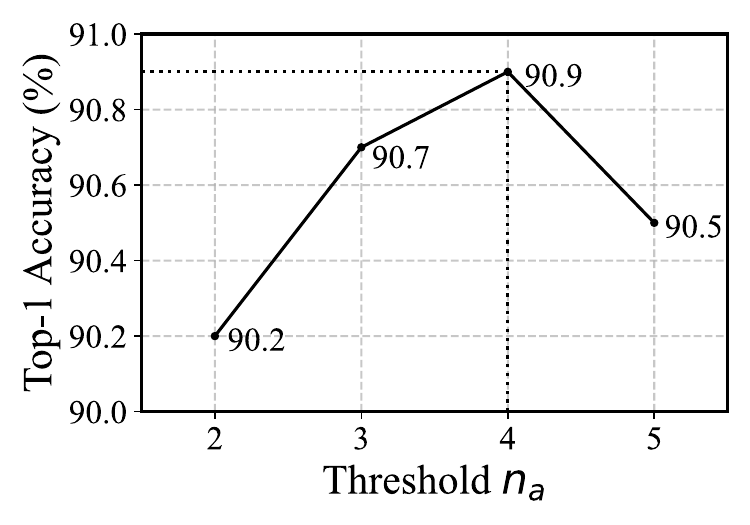}\label{fig_4_1}}
\hfill
\subfloat[Accuracy Vs. $\mathcal{L}_{inter}$ Weight ($\lambda_{1}$)]
{\includegraphics[width=.24\linewidth]{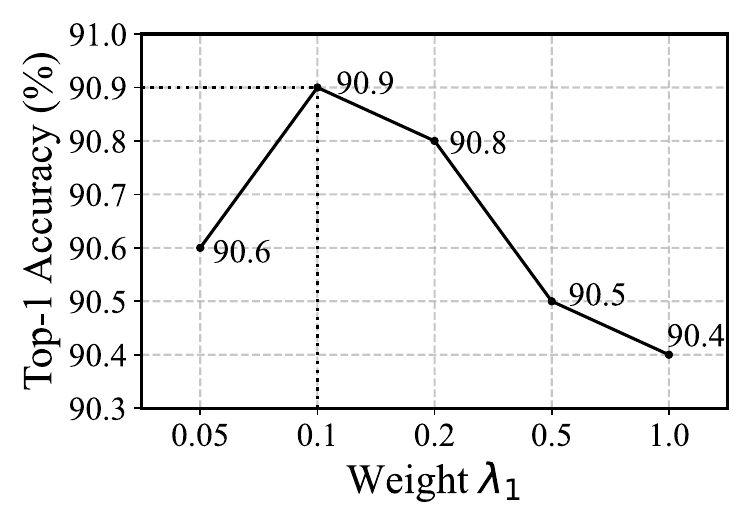}\label{fig_4_2}}
\hfill
\subfloat[Accuracy Vs. Margin ($\epsilon$)]
{\includegraphics[width=.24\linewidth]{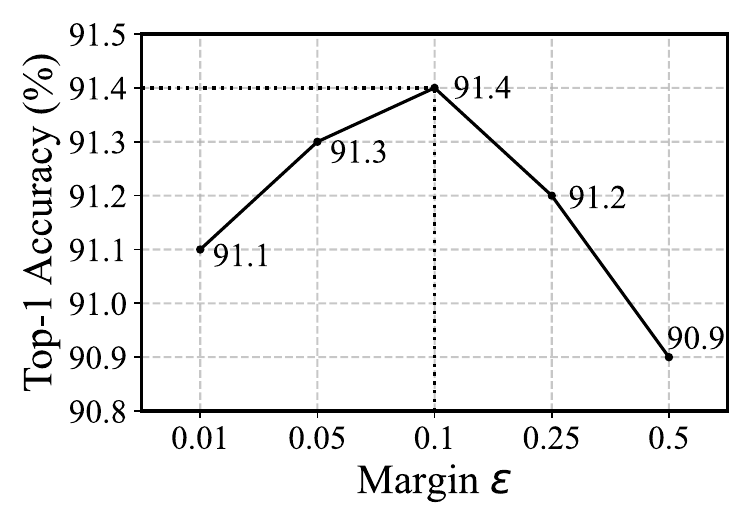}\label{fig_4_3}}
\hfill
\subfloat[Accuracy Vs. $\mathcal{L}_{intra}$ Weight ($\lambda_{2}$)]
{\includegraphics[width=.24\linewidth]{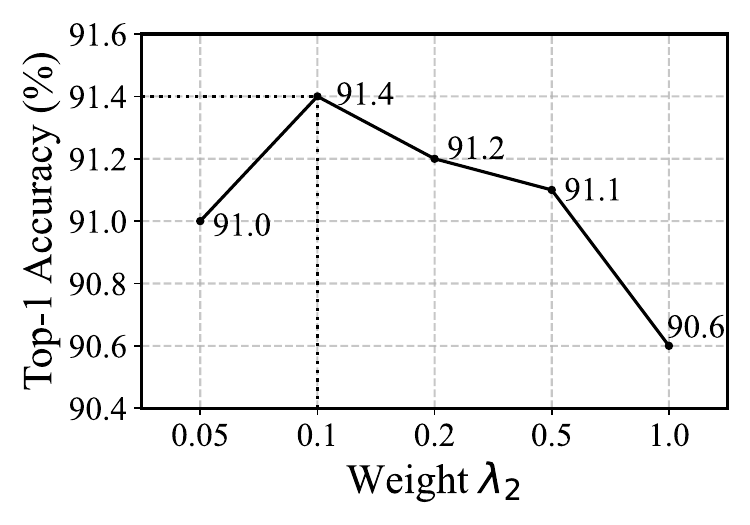}\label{fig_4_4}}
\caption{
Ablation study on the effect of different hyper-parameters under the NTU RGB+D 60 X-Sub setting with the joint modality.
}
\vspace{-0.5mm}
\label{fig:figure4}
\end{figure*}

\begin{table}[ht]
\centering
\caption{
Performance comparison of occlusions under the NTU-60 X-Sub setting in terms of accuracy (\%).
}
\renewcommand\arraystretch{1.18}
\resizebox{\linewidth}{!}{
\begin{tabular}{l c c c c c c}
\toprule
\multirow{2}{*}{Methods} & \multicolumn{5}{c}{Occluded Part} \cr
& Left Arm & Right Arm & Two Hands & Two Legs & Trunk \cr
\midrule
ST-GCN \cite{yan2018spatial} & 71.4 & 60.5 & 62.6 & 77.4 & 50.2 \\
2s-AGCN \cite{shi2019two} & 72.4 & 55.8 & 82.1 & 74.1 & 71.9 \\
RA-GCN \cite{song2020richly} & 74.5 & 59.4 & 74.2 & 83.2 & 72.3 \\
MS-G3D \cite{liu2020disentangling} & 31.3 & 23.8 & 17.1 & 78.3 & 61.6 \\
CTR-GCN \cite{chen2021channel} & 13.0 & 12.5 & 12.7 & 21.0 & 36.3 \\
HD-GCN \cite{lee2023hierarchically} & 67.1 & 55.7 & 56.7 & 74.8 & 61.3 \\
PDGCN \cite{chen2023occluded} & 76.0 & 62.0 & 75.4 & 85.0 & 73.0 \\
\midrule
Ours & \textbf{87.1} & \textbf{83.3} & \textbf{79.6} & \textbf{88.5} & \textbf{80.5}\\
\bottomrule
\end{tabular}
}
\label{tab:occlusion}
\end{table}

\subsubsection{Effectiveness of Individual Components}

We first scrutinize the contribution of each ACLNet component in Table~\ref{tab:ablation_1}.
Specifically, the introduction of $\mathcal{L}_{inter}$ (Inter-ACL) bolsters performance by 0.6\%.
We also provide the additional results that when incorporating the contextual similarity, the accuracy improves from 90.5\% with pairwise similarity alone to 90.9\%. 
This shows that the intricate inter-class relationships could provide valuable supervisory signals to help distinguish classes.
It is worth noting that the dynamic scheduling tailored for Inter-ACL acts primarily as hyper-parameter tuning when used alone, showing minimal impact on the baseline. 
Combining all the components, our full model reaches 91.4\% in accuracy and surpasses the baseline by 1.1\%.

\subsubsection{Implications of Motion Family}

In Fig.~\ref{fig:figure3}, we show the examples of Motion Family corresponding to the anchor actions `reading' and `wear jacket'.
We find that the constructed superclasses provide a good exploration of inter-class relationships between actions.
The method effectively captures the hidden, valuable connections and thus makes targeted distinctions.
Interestingly, it can be observed that within the Motion Family, some action classes are not intuitively related, and they appear to exhibit differences in motion patterns (e.g., reading and drinking water).
This is because the construction of the Motion Family reflects the similarity matrix statistics, highlighting shared spatial and temporal motion patterns among actions.
Upon analysis, we observe that actions like reading and drinking exhibit overlapping hand trajectories, particularly when picking up objects such as books or glasses.  
While these actions are clearly distinct in human vision, the model may confuse them based on the skeletal data. 
Grouping such related actions into one family helps to refine their differentiation. 
This targeted refinement enhances the ability of the model to handle harder cases.

\begin{figure*}[t!]
\centering
\subfloat[Epoch 10]
{\includegraphics[width=.24\linewidth]{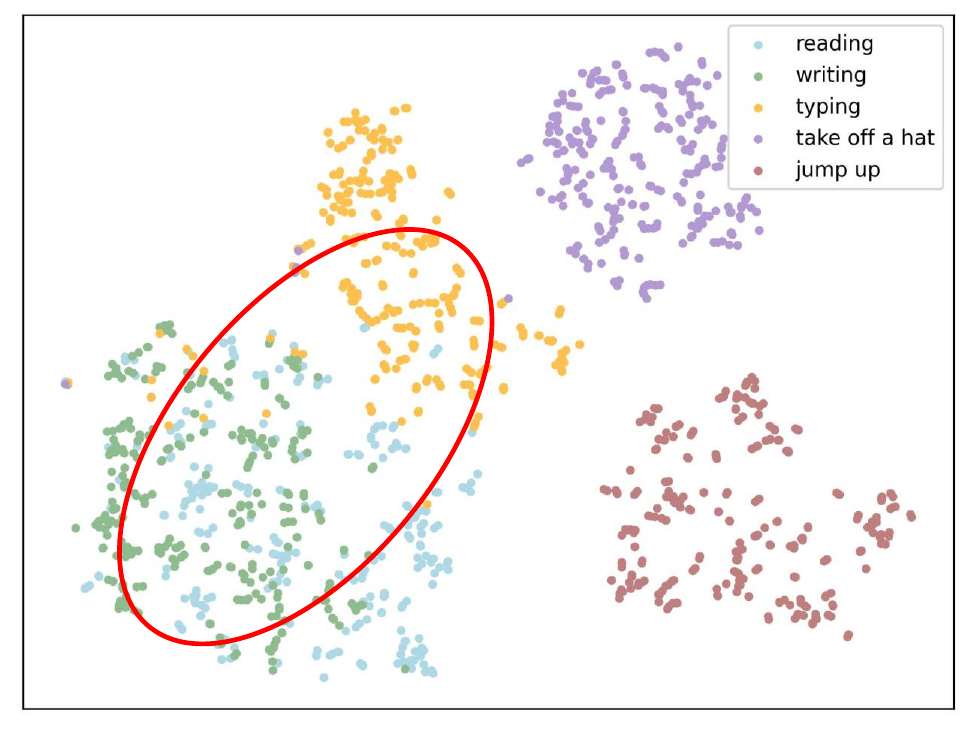}\label{fig_5_1}}
\hfill
\subfloat[Epoch 50]
{\includegraphics[width=.24\linewidth]{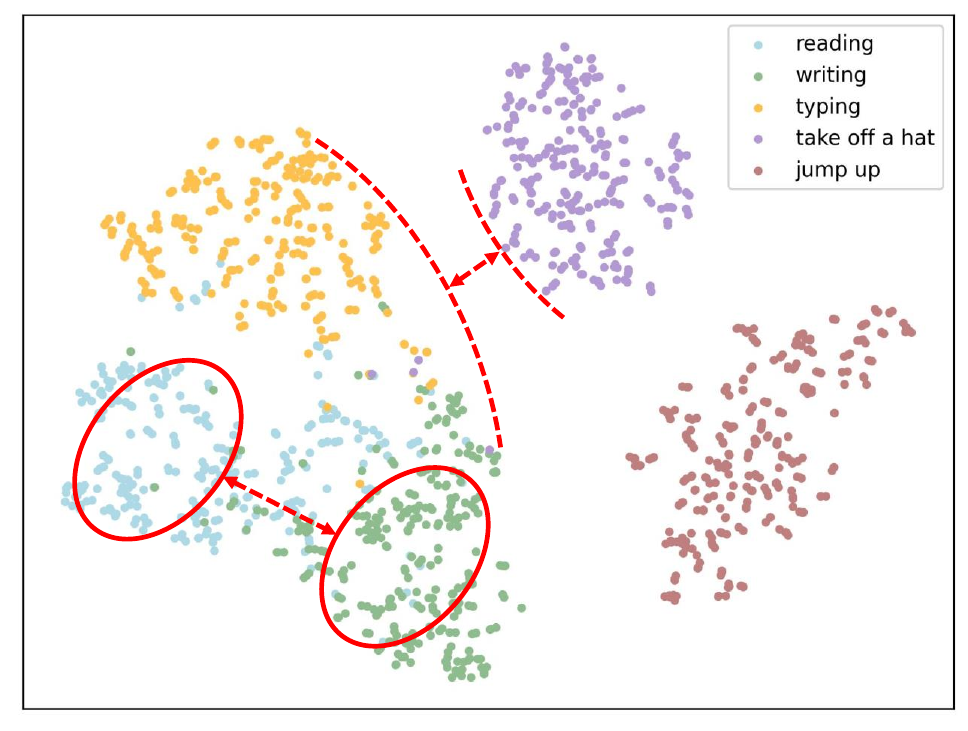}\label{fig_5_2}}
\hfill
\subfloat[Epoch 100]
{\includegraphics[width=.24\linewidth]{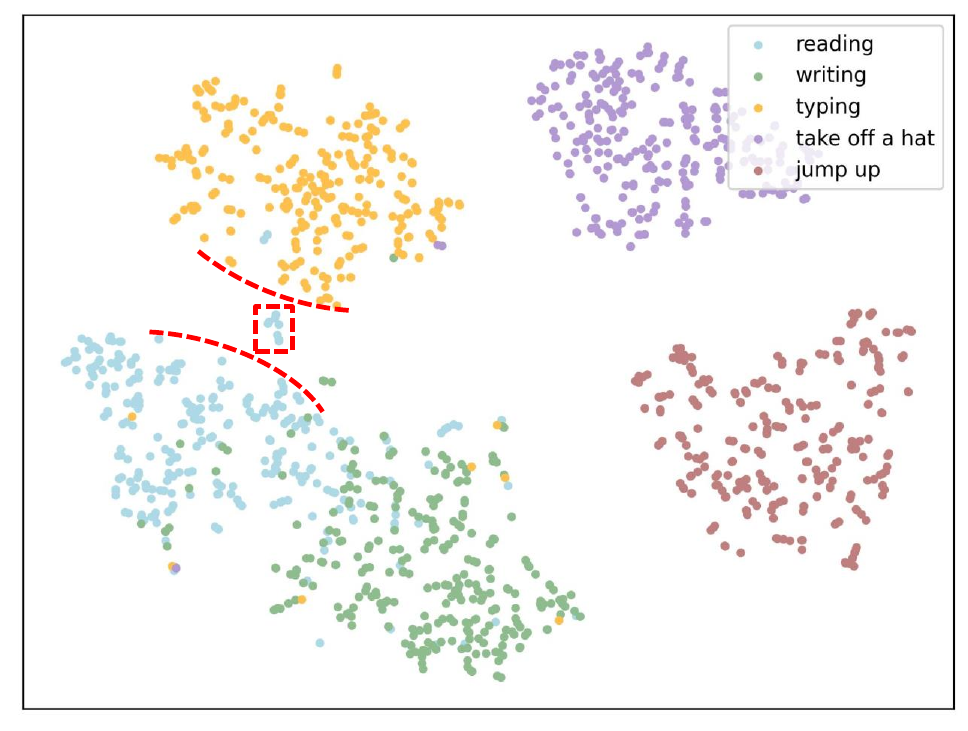}\label{fig_5_3}}
\hfill
\subfloat[Epoch 150]
{\includegraphics[width=.24\linewidth]{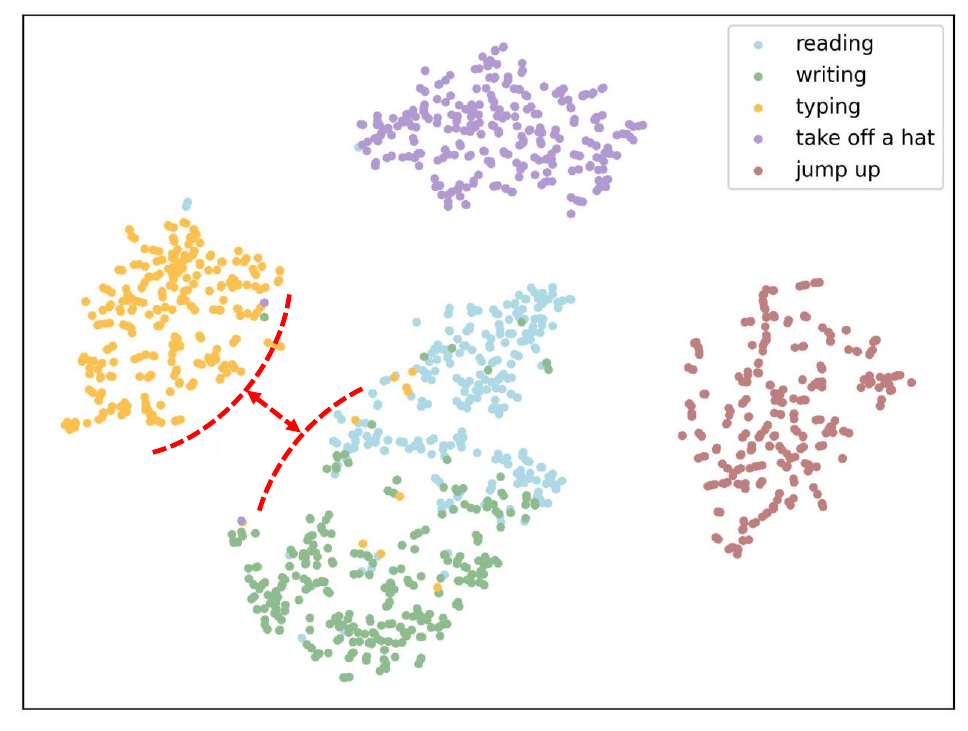}\label{fig_5_4}}
\caption{
The t-SNE plots of the feature embeddings for five chosen action classes throughout the training process. 
Colors indicate individual classes from NTU-60 X-Sub.
From the early epoch (a) to later epochs (b–d), the clusters grow progressively more compact and more widely separated, revealing a steady gain in class discriminability.
Best viewed with zoom in.
}
\vspace{-2mm}
\label{fig:figure5}
\end{figure*}

\begin{table}[t!]
\centering
\caption{ 
Average performance (\%) on different difficulty level classes sorted by accuracy under the NTU RGB+D 60 X-Sub setting.
}
\setlength\tabcolsep{12pt}
\renewcommand\arraystretch{1.1}
\begin{tabular}{l c c c}
\toprule
Difficulty Level  & Baseline & ACLNet & $\Delta$ \\
\midrule
class 1-5 & 68.9 & 72.0 & \textbf{+3.1} \\
class 6-15 & 82.2 & 84.8 & \textbf{+2.6} \\
class 16-30 & 90.7 & 92.1 & \textbf{+1.4} \\
class 31-60 & 96.3 & 96.6 & \textbf{+0.3} \\
\bottomrule
\end{tabular}
\label{tab:ablation_3}
\end{table}

\subsubsection{Effect of Hyper-parameters}

We analyze the effect of hyper-parameters in ACLNet, and the results are shown in Fig.~\ref{fig:figure4}.
The effect of the hyper-parameter $n_{a}$ is presented in Fig.~\ref{fig:figure4}~\subref{fig_4_1}. 
It can be observed that the appropriate threshold, \textit{i.e.}, $n_{a}=4$, benefits the classification performance. 
Then, we explore the impact of the weight $\lambda_{1}$ on $\mathcal{L}_{inter}$. 
As shown in Fig.~\ref{fig:figure4}~\subref{fig_4_2}, the best performance is achieved when $\lambda_{1}$ is equal to 0.1.
The impact of the margin $\epsilon$ is shown in Fig.~\ref{fig:figure4}~\subref{fig_4_3}.
We find that $\epsilon=0.1$ achieves the best performance.
Similarly, we examine the influences of $\lambda_{2}$ on $\mathcal{L}_{intra}$.
The results in Fig.~\ref{fig:figure4}~\subref{fig_4_4} show that $\lambda_{2}=0.1$ gives the best performance.

\subsubsection{Robustness to Noisy Skeleton Data}

Following the research~\cite{chen2023occluded}, we construct noisy skeleton data by simulating occlusions on the NTU-60 dataset. 
Specifically, we evaluate models using skeletons without joints of the left arm, right arm, two hands, two legs, and trunk, respectively. 
The models are trained on normal skeleton data and tested on incomplete skeleton data. 
As shown in Table~\ref{tab:occlusion}, our method achieves superior accuracy and demonstrates remarkable robustness.

\subsubsection{Visualization}

In Fig.~\ref{fig:figure5}, we plot the t-SNE visualization of skeleton representation distribution for five action classes.
As training progresses, it can be observed that the distinction between similar actions related to hand movements gradually becomes more prominent.
Moreover, the changes in Fig.~\ref{fig:figure5}~\subref{fig_5_3} and \subref{fig_5_4} demonstrate the effectiveness of the proposed method.
Hard samples between the two actions of `reading’ and `typing’ could be effectively distinguished, which further improves the separability within the classes.

\begin{figure}[t!]
\centering
\includegraphics[width=\linewidth]{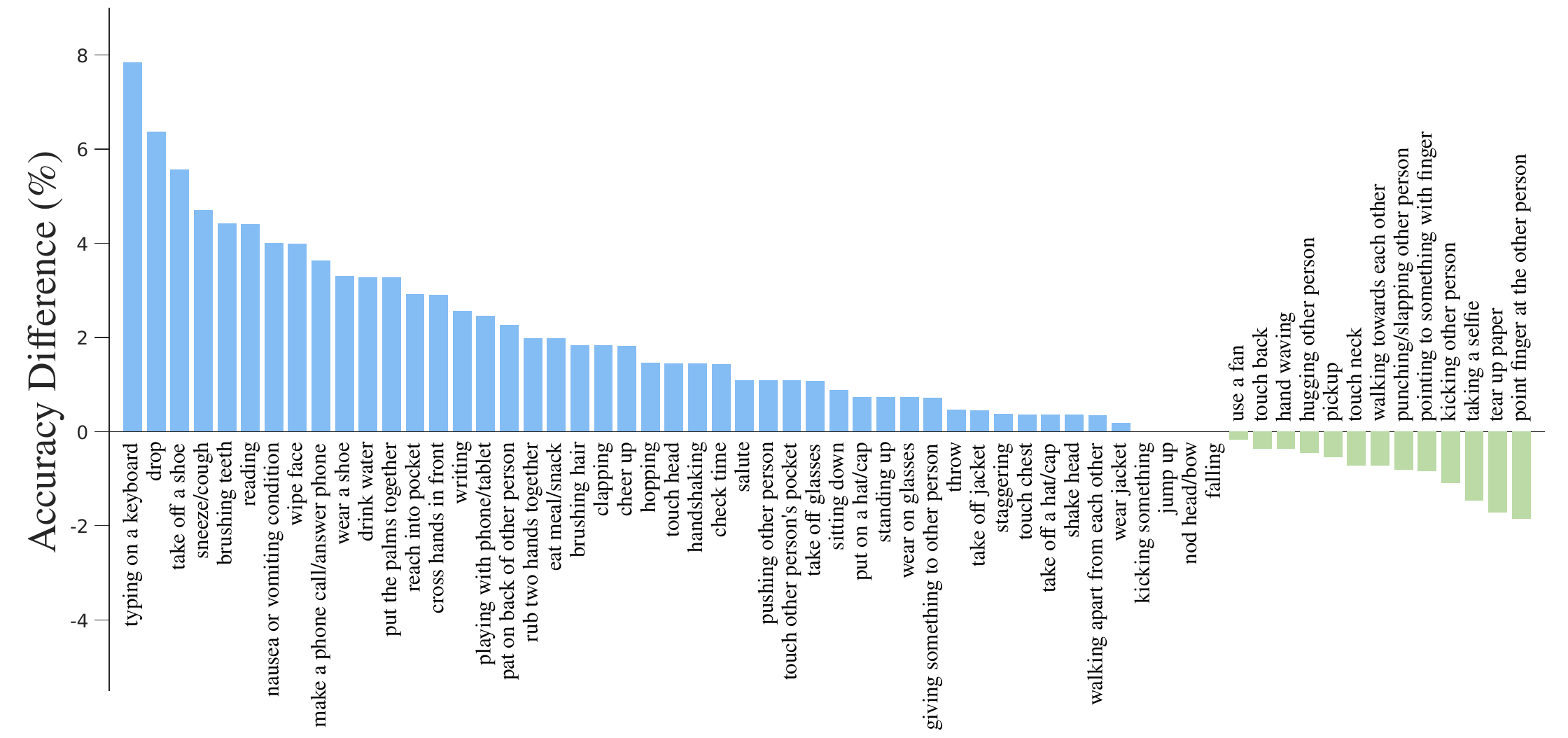}
\caption{ 
The accuracy difference (\%) between our method and FR-Head on NTU-60 X-Sub with the joint modality.
It is evident that ACLNet achieves more pronounced improvements in a greater number of classes.
}
\vspace{-2mm}
\label{fig:figure6}
\end{figure}

\subsubsection{Improvement on Similar Classes}

To evaluate the capability of the proposed method, we analyze the performance in discerning actions with differing degrees of similarity in Table~\ref{tab:ablation_3}. 
According to the results obtained with the baseline model, we sort 60 action classes of NTU RGB+D 60 based on classification accuracies, ranging from low to high. 
We categorize them into four difficulty levels and computed the average accuracy for each level. 
The results show that the proposed method achieves a more significant accuracy gain, especially in classes with higher difficulties.
Then, as shown in Fig.~\ref{fig:figure6}, we conduct the comparative evaluation of the class-wise accuracy difference.
It is evident that our method achieves more pronounced improvements in a significantly greater number of classes.
Additionally, for actions with temporal dependencies, we agree that explicitly incorporating finer-grained temporal dynamics into affinity modeling (e.g., phase-aware or segment-level affinities) is a promising direction.
We also acknowledge the potential of multimodal extensions that incorporate additional contextual information.
Moreover, the lack of necessary motion details, such as fingers for `point finger at the other person', would undermine the fine-grained differentiation.
Nevertheless, for similar actions, the recognition model could achieve superior performance.
These results demonstrate the effectiveness of the proposed method.

\section{Conclusion}

In this paper, we introduce ACLNet, a novel affinity contrastive learning network for skeleton-based human activity understanding. 
Concretely, our approach addresses the limitations in existing methods through two main contributions.
First, we introduce the concept of affinity similarity to model semantic relationships among hard classes, enabling targeted refinement through inter-class affinity learning. 
Second, we propose a marginal contrastive strategy that explicitly controls the separation between hard positives and negatives, enhancing robustness to intra-class variations.
Extensive experiments on six benchmarks demonstrate the effectiveness of ACLNet across skeleton-based action recognition, gait recognition, and person re-identification tasks. 
The proposed affinity modeling paradigm opens new avenues for fine-grained activity analysis and behavioral biometrics, with potential applications in security, healthcare, and human-computer interaction.

\bibliographystyle{IEEEtran}
\bibliography{trans}

\end{document}